\documentclass[runningheads,a4paper]{llncs}

\usepackage{amssymb}
\usepackage{graphicx}
\usepackage{pgfplots}
\usepackage{pgfplotstable}
\usepackage{multirow}
\usepackage{tikz}
\usetikzlibrary{arrows}
\usepackage{amsmath}
\usepackage{makeidx}
\usepackage{caption}
\usepackage{subcaption}
\usepackage{subfig} 
\usepackage{color}
\usepackage[normalem]{ulem}
\usepackage{url}
\usepackage{mathtools}
\usepackage{verbatim}

\newcommand{\keywords}[1]{\par\addvspace\baselineskip
\noindent\keywordname\enspace\ignorespaces#1}

\begin{document}

\mainmatter  
\title{A Subpath Kernel for Learning Hierarchical Image Representations}
\titlerunning{A Subpath Kernel for Learning Hierarchical Image Representations}

%
%
\author{Yanwei Cui\and Laetitia Chapel\and S\'ebastien Lef{\`e}vre}
\authorrunning{A Subpath Kernel for Learning Hierarchical Image Representations}

\institute{ IRISA (UMR 6074), Univ. Bretagne-Sud, Vannes, France}

\maketitle

\begin{abstract}
Tree kernels have demonstrated their ability to deal with hierarchical data, as the intrinsic tree structure often plays a discriminative role. While such kernels have been successfully applied to various domains such as nature language processing and bioinformatics, they mostly concentrate on ordered trees and whose nodes are described by symbolic data.  
Meanwhile, hierarchical representations have gained increasing interest to describe image content. This is particularly true in remote sensing, where such representations allow for revealing different objects of interest at various scales through a tree structure. However, the induced trees are unordered and the nodes are equipped with numerical features.
In this paper, we propose a new structured kernel for hierarchical image representations which is built on the concept of subpath kernel. Experimental results on both artificial and remote sensing datasets show that the proposed kernel manages to deal with the hierarchical nature of the data, leading to better classification rates. \footnote{10th IAPR-TC15 Workshop on Graph-based Representations in Pattern Recognition (GbR2015) in NLPR of Institute of Automation, Chinese Academy of Sciences. Beijing, China
}

\keywords{Hierarchical representation, image classification, structured kernels, subpath kernel}
\end{abstract}

\section{Introduction}

Structured data-based learning has become a central topic in machine learning, as such data representations are met in numerous fields. We focus here on tree-based representations, whose typical applications are parse trees in Natural Language Processing \cite{collins2001convolution},  XML trees in web mining \cite{kimura2011subpath}, or even hierarchical image representations \cite{blaschke2014geographic}. 
Since tree structure plays an important role in tasks like classification or clustering, similarity measures taking explicitly into account topological characteristics of the tree are sought. Among them, kernels functions are appealing as they allow the use of popular kernel methods \cite{Shawe:2004}. 

Various kernels have indeed been proposed to cope with a tree as the underlying data structure (see \cite{Yukiko2014} for a review). They mostly rely on a fundamental idea brought by Haussler with its convolution kernels \cite{haussler1999convolution}, stating that a kernel defined on a complex structure can be formed by kernels computed on its substructures. Most often, those kernels are defined for ordered trees, that is to say trees for which left to right order among nodes or leaves is fixed (mostly because of the specific nature of the data or due to computational complexity issues). Examples of such kernels include the subset tree kernel \cite{collins2001convolution} and the subtree kernel \cite{vishwanathan2004fast}. Unordered trees received much less attention, with the subpath kernel \cite{kimura2011subpath} being one of the very few existing solutions.

Meanwhile, an emerging paradigm for image classification has advocated the idea of relying on hierarchical representations \cite{blaschke2014geographic}, which are  built using series of nested partitions or segmentations, rather than the usual flat representation. This is particularly true in remote sensing, where such representations allow for revealing different objects of interest at various scales through a tree structure. However, the induced trees are unordered and the nodes or regions are associated with numerical features, preventing the use of existing tree kernels. 

We propose in this paper a new kernel that arises from the subpath kernel \cite{kimura2011subpath}. Based on some existing adaptations to numerical data from the graph kernel literature, the designed kernel is able to cope with unordered trees equipped with numerical data (see Sec.~\ref{Subpath kernel}). Besides, it considers the complete set of subpaths on tree structures (instead of paths on graphs), leading to an efficient computation scheme. Experimental results are given in Sec.~\ref{Experiments}. They rely on artificial datasets as well as a real multispectral satellite image. We end the paper with some concluding remarks and directions for future work.

\section{Proposed Kernel}\label{Subpath kernel}

We focus here on structured data represented by trees and subpaths. Let us first recall that a tree is a directed and connected acyclic graph with a single root node. A path connecting a node in the tree to one of its descendants is called a subpath. Individual nodes are also included in the set of subpaths.

We build upon the subpath kernel \cite{kimura2011subpath} to design a new kernel able to cope with numerical data. Let us recall the principles of the original subpath kernel, that exploits the hierarchical structure by counting all possible common subpaths embedded in two tree structures equipped with symbolic features. Given two trees $T$ and $T'$, the subpath kernel is defined as
\begin{equation} 
K(T,T')=\sum_{s \in T, s' \in T' } k(s,s' ) = \sum_{s \in T, s' \in T' } \delta_{s,s'}  \ h(s,T) \ h(s',T' ) \;  ,
\label{subpathSym}
\end{equation}
where kernels $k(s,s')$ are computed between all subpaths $s, s'$ of tree $T,T'$ respectively. They rely on the number of occurrences of the subpaths in the tree written $h(s,T)$ and $h(s',T')$. The Kronecker delta function $\delta_{s,s'}$ equals $1$ iif the two subpaths $s,s'$ are identical. Figure~\ref{fig:subpath} illustrates for a given simple tree its possible subpaths and their occurrences in the tree.

 \tikzset{
  treenode/.style = {align=center, inner sep=0pt, text centered,
    font=\sffamily},
  arn_n/.style = {treenode, circle, white, font=\sffamily\bfseries, draw=black,
    fill=black, text width=1.5em},
  arn_b/.style = {treenode, circle, black, draw=black, 
    text width=1.5em},
  arn_g/.style = {treenode, circle, gray, draw=gray,
    fill=gray,text width=1.5em}
}
\begin{figure}[ht]
\centering
\begin{subfigure}[t]{0.36\textwidth}
\centering
\begin{tikzpicture}[->,>=stealth',level/.style={sibling distance = 2cm/#1,
  level distance = 1cm}] 
\node [arn_b] {A}
    child{ node [arn_b] {B}}
    child{ node [arn_b] {B}
            child{ node [arn_b] {C} 
            }
            	}
;
\end{tikzpicture}
\caption{Tree $T$}
\end{subfigure}
\centering
\begin{subfigure}[t]{0.6\textwidth}
\centering
\begin{tikzpicture}[->,>=stealth',level/.style={sibling distance = 2cm/#1,
  level distance = 1cm}] 
\node [arn_b] {A}
    child{ node [arn_b] {B}
            child{ node [arn_b] {C} 
            }
            	}
;
\end{tikzpicture}
\begin{tikzpicture}[->,>=stealth',level/.style={sibling distance = 2cm/#1,
  level distance = 1cm}] 
\node [arn_b] {A}
    child{ node [arn_b] {B}}
;
\end{tikzpicture}
\begin{tikzpicture}[->,>=stealth',level/.style={sibling distance = 2cm/#1,
  level distance = 1cm}] 
\node [arn_b] {B}
    child{ node [arn_b] {C}}
;
\end{tikzpicture}
\begin{tikzpicture}[->,>=stealth',level/.style={sibling distance = 2cm/#1,
  level distance = 1cm}] 
\node [arn_b] {A}
;
\end{tikzpicture}
\begin{tikzpicture}[->,>=stealth',level/.style={sibling distance = 2cm/#1,
  level distance = 1cm}] 
\node [arn_b] {B}
;
\end{tikzpicture}
\begin{tikzpicture}[->,>=stealth',level/.style={sibling distance = 2cm/#1,
  level distance = 1cm}] 
\node [arn_b] {C}
;

\end{tikzpicture}
\caption{Enumerate all subpaths $s_i \in T$ }
\end{subfigure}
\caption{A tree $T$ and all its subpaths $s_i$, that may have multiple occurrences in $T$: $h(s_i,T)=2$ for $s_{2}=(A \rightarrow B)$ and $s_{5} = (B)$ ; $h(s_i,T)=1$ otherwise.}
\label{fig:subpath}
\end{figure}
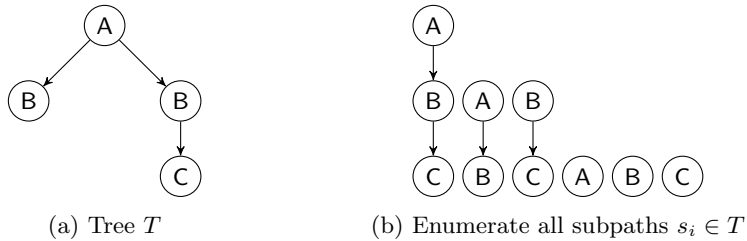

\subsection{Adaptation to numeric data}
The original kernel depicted previously was introduced for data classification in bioinformatics where nodes take symbolic values. Adapting this kernel to numeric data requires one to change the terms $h(s,T)$ and $\delta_{s,s'}$ since strict identity between subpaths (and their respective node features) does not generally occur. We follow here the scheme proposed with graph kernels for image classification \cite{aldea2007image}, but considering subpaths instead of random walks as the substructure component. Indeed, the use of trees allows a complete enumeration of the subpaths that is not achievable with graphs. 

We replace $h(s, T)$ and $h(s',T')$ terms in Eq.~\eqref{subpathSym} by a product of some atomic kernel functions $k(n_i,n'_i)$ computed between pairs of nodes $n_i \in s$ and $n'_i \in s'$.  Various atomic kernels are here available, \emph{e.g.} Gaussian kernel \cite{Shawe:2004} that has often been successfully used in many contexts. As long as these atomic kernels are positive definite, the proposed structured kernel is also positive definite (see \cite{haussler1999convolution}). 
Formulation of the subpath kernel for numeric data is then
\begin{equation} 
K(T,T')=\sum_{s \in T, s' \in T' } k(s,s' ) = \sum_{s \in T, s' \in T' } \delta_{|s|,|s'|} \  \prod_{\substack{n_i \in s, n'_i \in s'}} k(n_i,n'_i)\;  ,
\label{subpathNum}
\end{equation}
where the Kronecker delta function $\delta_{|s|,|s'|}$ equals $1$ iif the two subpaths $s,s'$ have the same length, and nodes $n_i,n'_i$ are scanned in descending order along the two subpaths $s,s'$ (from the root to the leaf). One might notice that if $k(n_i,n'_i)=\delta_{n_i,n'_i}$ measures the identicalness of $n_i,n'_i$, Eq.~\eqref{subpathNum} becomes just another form of Eq.~\eqref{subpathSym}. The naive complexity of comparing all pairs of subpaths between $T$ and $T'$ is $O(|T|^2 |T'|^2)$ as indicated in \cite{kimura2011subpath} (the size of subpath set of a given tree T is in general $|T|^2$, with $|T|$ refers to the number of vertices in the tree).

\subsection{Efficient computation}
Besides the proposed adaptation to numeric data, applying kernels on tree-based representations of images also raises a computational issue. Indeed, images are most often made of millions or even billions of pixels, that are put in the tree structure. While such structure aims to reduce the image content through a hierarchical representation, it may still be characterized by a huge number of nodes and edges. So we also need to address this issue to make the proposed kernel relevant when dealing with images.

Let us note that in the original subpath kernel paper \cite{kimura2011subpath}, some solutions were given to lower the computation time. But they are related to symbolic data and thus cannot be applied here. Inspired from \cite{harchaoui2007image}, we suggest rather to use dynamic programming for comparing all subpaths. This strategy allows us to break down the complexity of all subpaths comparison into smaller subproblems in a recursive way, and reuse the solution of one subproblem to solve another one. Repeated comparisons are then avoided. More specifically, the comparison is computed recursively between two nodes $n \in T$ and $n' \in T'$: 
\begin{equation}
k^p(n,n')  = k(n,n')+ \Bigl(k(n,n')+1\Bigr)\sum_{\mathclap{\substack{n_c \in C_T(n), n'_c \in C_{T'}(n')}}} k^{p-1}(n_c,n'_c)
- k(n,n') \sum_{\mathclap{\substack{n_c \in C_T(n), n'_c \in C_{T'}(n')}}}k^{p-2}(n_c,n'_c) \;  ,
\end{equation}
with $C_T(n)$ the set of children of $n$ in $T$, and $k^1(n,n')= k(n,n') $ the atomic kernel measuring the similarity between $n$ and $n'$, $k^0(n,n')= 0$ by convention. We also have $p$ reaching 1 when either $n$ or $n'$ has no child. For more details, let us denote $ST_n,ST_{n'}$ two subtrees rooted at $n \in T$, $n' \in T'$ respectively, $(s,s')$ a pair of subpaths with $s \in ST_n, s' \in ST_{n'} $. Then $k^p(n,n')$ sums the similarity measures of all pairs of subpaths $(s,s')$ with same length and starting at the same height in $ST_n,ST_{n'}$. The similarity is calculated recursively by the first and second term on the right side of equation, together with the third term $k^{p-2}$ introduced to prevent false subpaths that compare non-contiguous alignments.

Given a couple of trees $T$ and $T'$ with respective roots $r(T)$ and $r(T')$, we finally compute all pairs of subpaths embedded in the trees:
\begin{equation}
K(T,T')= k^p(r(T),r(T'))  + \sum_{\mathclap{n' \in T', n' \neq r(T') }} k^p(r(T),n')
+ \sum_{\mathclap{n \in T, n \neq r(T)}} k^p(n,r(T')) \;  .
\end{equation}

Dynamic programming allows us to avoid the explicit computation of all pairs of subpaths between two trees $T_1$ and $T_2$. Instead, only all pairs of vertices are considered. The complexity is thus $O(|T_1||T_2|)$ where $|T_i|$ refers to the number of vertices in the tree $T_i$.

\subsection{Additional improvements}
The proposed kernel shares with existing structured kernels two main issues. On the one hand, the value of $K(T,T')$ depends on the size of the trees while some invariance might be sought. This problem has already been tackled in  \cite{collins2001convolution} through a normalized kernel, that is computed here as 
\begin{equation} 
K(T,T')=K(T,T') \ \bigl( K(T,T)K(T',T') \bigr)^{-0.5}.
\end{equation}
On the other hand, the structured kernel gives the same importance to every node in the tree. Here nodes represent regions of various size, and larger regions might be given more attention than smaller ones. By weighting the atomic kernel by the relative size $A_n$ (\emph{i.e.}, number of pixels) of a node w.r.t. the root of the tree, and given a parameter $\beta \geq 0$, the updated kernel becomes
\begin{equation} 
A_n^\beta A_{n'}^\beta k(n,n') \;  .
\label{atomic}
\end{equation}

\section{Experiments} \label{Experiments}

We have conducted two experiments to proceed to a finer understanding of the kernel behavior using an artificial dataset, and to validate the kernel in a realistic context. Before giving in-depth analysis of the results obtained on both datasets, we will present first the common experimental setup.

\subsection{Experimental Setup}

We evaluate kernels in a classification context, considering a \textit{one-against-one} SVM classifier (using the Java implementation of LibSVM \cite{chang2011}). The proposed subpath kernel is systematically compared with a kernel computed on the root of the tree only (\emph{i.e.} ignoring all remaining nodes of the tree), called rooted kernel in the sequel. Our goal is to assess the importance of the various levels of information contained in the hierarchical representation w.r.t. a raw analysis of the whole data. Let us note that  standard tree kernels based on substructures other than paths could hardly be applied here due to their computational complexity \cite{Yukiko2014}). For the proposed kernel, nodes are compared individually using an atomic RBF kernel: each node being described by a feature or vector of $N$ attributes, the kernel is defined for a pair of nodes $n,n'$ with respective features $x,x'$ as 
\begin{equation} 
 k(n,n') = \exp (-\gamma d(x,x'))\;.
\label{rbf}
\end{equation}
We consider here two types of distances. The first one is an $l^2\textendash norm$ distance 
\begin{equation}
d_\mathcal{G}(x,x')=\|x- x' \|^2 \; ,
\end{equation}
leading to a Gaussian kernel, for which the features $x,x'$ contain the average and variance computed from all information contained in the nodes (that can be accessed from their leaves). The second one is a distance between $N$-d histograms:
\begin{equation} 
d_{\chi^2}(x,x') = \sum_{j=1}^{N} \frac{(x_j - x'_j)^2}{x_j + x'_j} \; ,
\end{equation}
where the features $x,x'$ are here histograms of $M$ bins per dimension stacked together. We call $\chi^{2}$ kernel the resulting atomic kernel.

Three free parameters are determined by a grid search strategy over potential values: the bandwidth $\gamma$  of the RBF atomic kernel (Eq. \eqref{rbf}), the SVM regularization parameter $C$, and the size weight $\beta\in[0,1]$ (Eq. \eqref{atomic}).

Accuracies (and standard deviations) of each setup are computed after 100 repetitions of each experiment, choosing randomly 20 data samples from each class as training samples,  using the remaining samples for testing. 

\subsection{Artificial dataset}
In this first experiment, we study the behavior of the proposed tree kernel though 3 different scenarios using an artificial dataset. Unless stated otherwise, we consider $M = 4$ bins by dimension to construct the histogram. We  call structure information the way leaves are aggregated and the initial number of leaves. 

\tikzset{
  treenode/.style = {align=center, inner sep=0pt, text centered,
    font=\sffamily},
  arn_n/.style = {treenode, circle, white, font=\sffamily\bfseries, draw=black,
    fill=black, text width=1em},
  arn_b/.style = {treenode, circle, black, draw=black, 
    text width=1em},
  arn_g/.style = {treenode, circle, gray, draw=gray,
    fill=gray,text width=1em}
}
\begin{figure}[!h]
        \centering
        \begin{subfigure}[t]{0.3\textwidth}
                \centering
                \begin{tikzpicture}[->,>=stealth',level/.style={sibling distance = 1.6cm/#1,
                  level distance = 0.7cm}] 
                 \node at (0,0.5) (node) {Class 1};      
                \node [arn_n] {}
                    child{ node [arn_n] {}
                            child{ node [arn_n] {A} 
                            }
                            child{ node [arn_n] {A}
                            }
                	}
                    child{ node [arn_n] {}
                            child{ node [arn_n] {A} 
                            }
                            child{ node [arn_n] {A}
                            }
                	}
                ;
                \end{tikzpicture}
                
                ~
                
                \begin{tikzpicture}[->,>=stealth',level/.style={sibling distance = 1.6cm/#1,
                  level distance = 0.7cm}] 
                \node at (0,0.5) (node) {Class 2};      
                \node [arn_b] {}
                    child{ node [arn_b] {} 
                            child{ node [arn_b] {B} 
                                    	 }
                            child{ node [arn_b] {B}				
                            }                            
                    }
                    child{ node [arn_b] {}
                            child{ node [arn_b] {B} 
                                    	 }
                            child{ node [arn_b] {B}				
                            }  
                }
                ;
                \end{tikzpicture}
                \caption{non discriminative tree, discriminative root and nodes.}
                \label{fig:toyCase1}
        \end{subfigure}%
        ~ 
        \begin{subfigure}[t]{0.3\textwidth}
                \centering
                \begin{tikzpicture}[->,>=stealth',level/.style={sibling distance = 1.6cm/#1,
                  level distance = 0.7cm}] 
                 \node at (0,0.5) (node) {Class 1};      
                \node [arn_n] {}
                    child{ node [arn_n] {}
                            child{ node [arn_n] {A} 
                            }
                            child{ node [arn_n] {A}
                            }
                	}
                    child{ node [arn_n] {}
                            child{ node [arn_n] {A} 
                            }
                            child{ node [arn_n] {A}
                            }
                	}
                ;
                \end{tikzpicture}        	
     
                ~
                
                \begin{tikzpicture}[->,>=stealth',level/.style={sibling distance = 0.8cm/#1,
                  level distance = 1.4cm}] 
                 \node at (0,0.5) (node) {Class 2};      
                \node [arn_n] {}
                    child{ node [arn_n] {A} }
                    child{ node [arn_n] {A} }
					child{ node [arn_n] {A} }	
					child{ node [arn_n] {A} }	                                               
                    
                ;
                \end{tikzpicture}
                
                \caption{non discriminative root and nodes, discriminative tree.}
                \label{fig:toyCase2}
        \end{subfigure}
        ~ 
        \begin{subfigure}[t]{0.3\textwidth}
        \centering
                \begin{tikzpicture}[->,>=stealth',level/.style={sibling distance = 1.6cm/#1,
                  level distance = 0.7cm}] 
                 \node at (0,0.5) (node) {Class 1};      
                \node [arn_g] {}
                    child{ node [arn_g] {} 
                            child{ node [arn_n] {A} 
                            	 }
                            child{ node [arn_b] {B}				
                            }                            
                    }
                    child{ node [arn_g] {}
                            child{ node [arn_n] {A} 
                            	 }
                            child{ node [arn_b] {B}				
                            }
                }
                ;
                \end{tikzpicture}

                ~
                                
                \begin{tikzpicture}[->,>=stealth',level/.style={sibling distance = 1.6cm/#1,
                  level distance = 0.7cm}] 
                 \node at (0,0.5) (node) {Class 2};      
                \node [arn_g] {}
                    child{ node [arn_n] {} 
                            child{ node [arn_n] {A} 
                            }
                            child{ node [arn_n] {A}
                            }                         
                    }
                    child{ node [arn_b] {}
                            child{ node [arn_b] {B} 
                                    	 }
                            child{ node [arn_b] {B}				
                            }  
                }
                ;
                \end{tikzpicture}
                \caption{non discriminative root and tree, discriminative nodes.}
                \label{fig:toyCase3}
        \end{subfigure}
        \caption{An illustration of the different scenarios used for experimental evaluation.}\label{fig:toyset}
\end{figure}
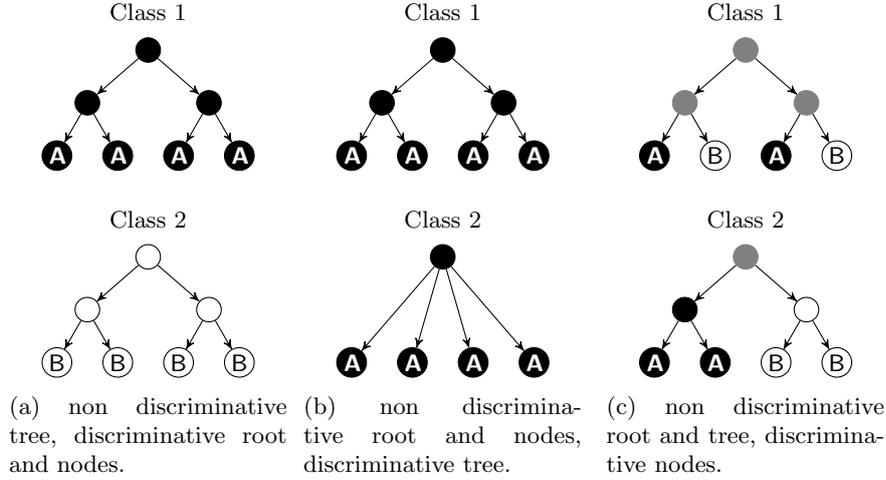

\textbf{(a) Only the root is discriminative.} We generate two types of leaves, $A$ and $B$, that are described by a 1-D feature generated according to a uniform distribution, with non overlapping intervals. \texttt{Class 1} trees are composed of leaves of type $A$ only and \texttt{Class 2} trees of type $B$ only (see Fig.~\ref{fig:toyCase1}). Number of leaves and node merging parameters are defined randomly to produce various shapes of trees within each class. As shown in Tab.~\ref{tab:toy}a, subpath kernel behaves similarly to rooted kernel: when the structure does not provide additional information, exploiting it in the proposed subpath kernel does not degrade the performances.

\textbf{(b) Only the structure is discriminative.} We generate only type $A$ leaves. The two classes of trees can then be discriminated thanks to their structure, \emph{i.e.} with different ranges of related parameters (number of leaves and number of fanouts for each node) for each class, see Fig.~\ref{fig:toyCase2}. As shown in Tab.~\ref{tab:toy}b, rooted kernel achieves only 50\% accuracy, while the subpath kernel is able to discriminate the two classes, thanks to the discriminative structure leading to different subpaths between the two classes. Let us note that when $\gamma=0$ and $\beta=0$, the subpath kernel turns into a kernel that computes the product of the number subpaths with common length embedded in the two trees.

\textbf{(c) Only the features of the nodes are discriminative.} We generate both type $A$ and $B$ leaves, and we force type $A$ leaves to merge with type $B$ leaves in \texttt{Class 1}, while in \texttt{Class 2}, type $A$ (resp. $B$) leaves always merge with type $A$ (resp. $B$) leaves (see Fig.~\ref{fig:toyCase3}). Similarly to scenario (a), structure parameters are selected randomly. As shown in Tab.~\ref{tab:toy}c, rooted kernel provides an accuracy about 50\%, due to the non discriminative root. Discriminative information contained in the nodes benefits to the proposed subpath kernel, leading to a 100\% accuracy. Note that even in the presence of irrelevant features, the subpath kernel still behaves correctly. Indeed, we have experimentally observed that adding 40 non discriminative features to each node (so only one dimension among the 41 is relevant) leads to an accuracy of 97.13 \% with Gaussian atomic kernel, and 99.99 \% with $\chi^{2}$ atomic kernel.

\textbf{(c1) Robustness to outliers.} We modify the scenario (c) to introduce outlier leaves that take values outside ranges of type $A$ and type $B$ leaves. The ratio of such leaves varies from 0\% to 100\%. We construct here (and here only) the histogram for the $\chi^{2}$ atomic kernel considering $M=12$ bins. 

\textbf{(c2) Robustness to mislabelled leaves.} We update the scenario (c1) with outlier leaves changed to mislabelled leaves. To do so, some leaves of type $A$ are changed into type $B$, and vice versa. In the binary classification setup considered here, the ratio of mislabelled leaves in each class varies from 0 \% to 50 \%, leading to more confusing subpaths between the two classes.

\begin{figure}[!ht]
        \centering
        \begin{subfigure}[t]{0.48\textwidth}     
				\includegraphics[width=6cm,height=3cm]{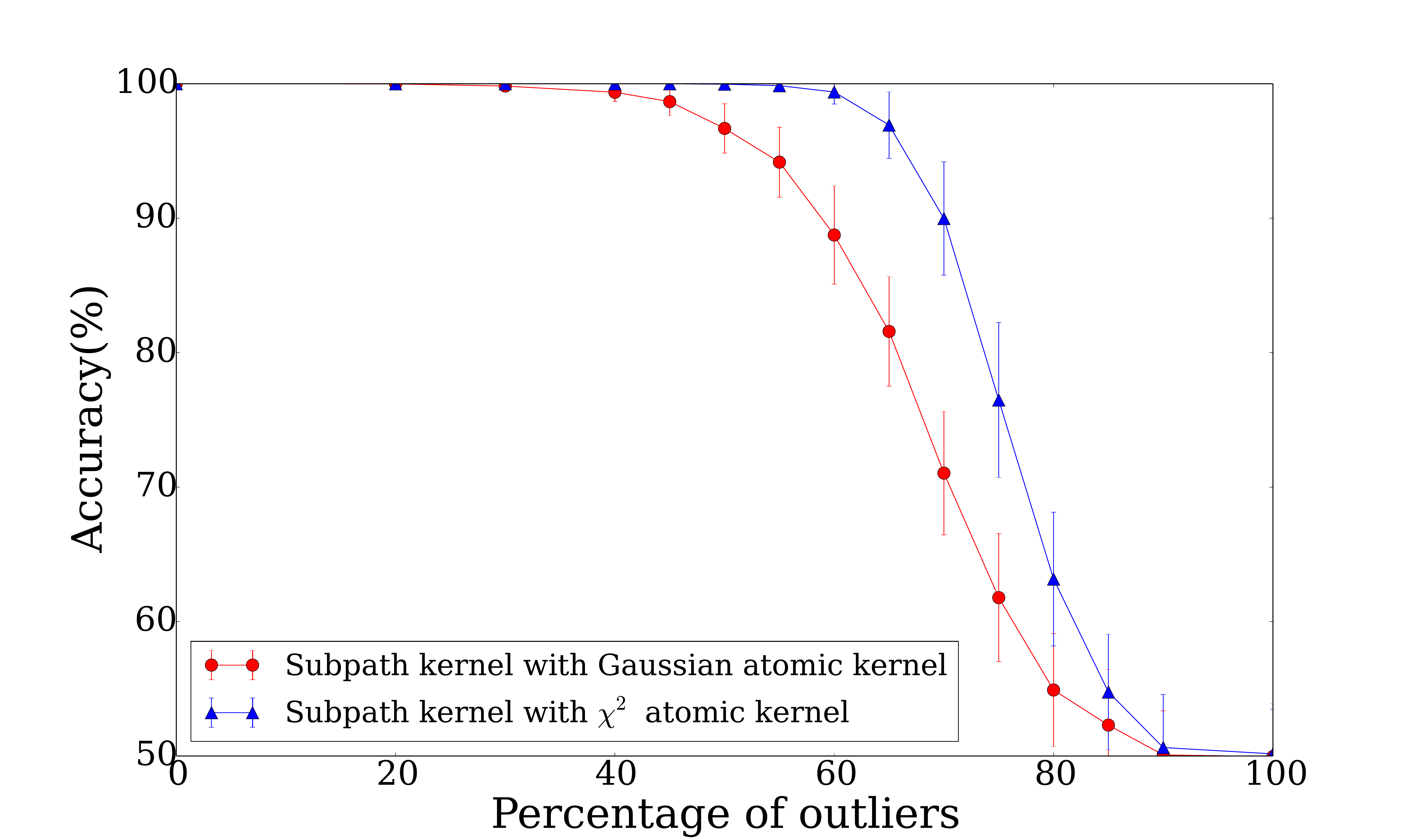}
                \caption{Robustness to outliers.}
                \label{fig:resCase3.1}
        \end{subfigure}
        \begin{subfigure}[t]{0.48\textwidth}
        		\includegraphics[width=6cm,height=3cm]{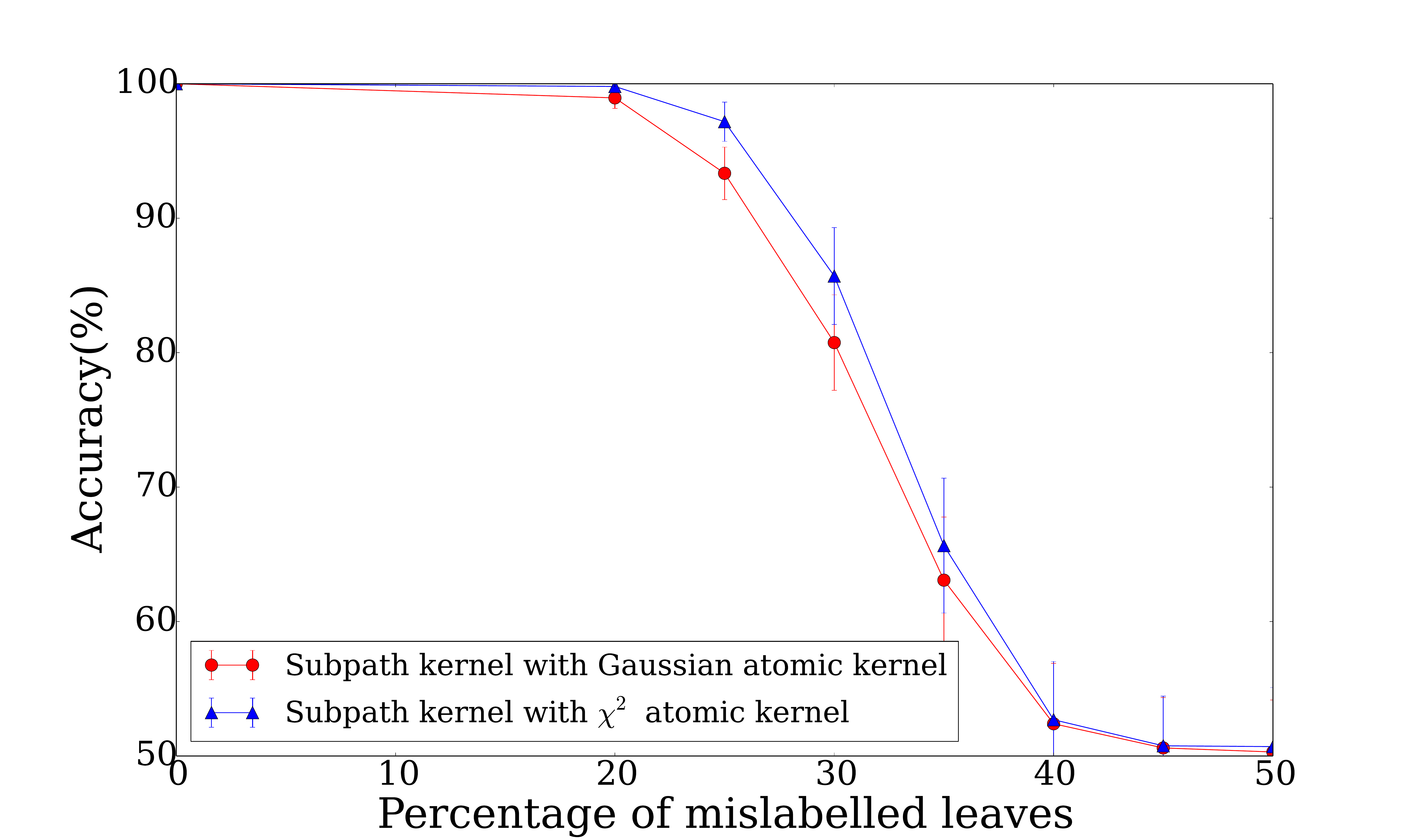}
                \caption{Robustness to mislabelled leaves.}
                \label{fig:resCase3.2}
        \end{subfigure}
        \caption{Accuracies for scenarios \textbf{(c1)} and \textbf{(c2)}.}
        \label{fig:robustness}
\end{figure}

We can derive two main observations from Fig.~\ref{fig:robustness}: a) the subpath kernel can maintain a good performance up to a certain ratio of structure distortion, and b) both accuracy drops (after 50\% of outliers in \textbf{c1}, between 20\% and 40\% mislabelled leaves in \textbf{c2}) illustrate that subpath kernel performance is directly related to the discrimination of substructures between two group of complex structured data.
Further, one might notice that in both scenarios, subpath using $\chi^{2}$ atomic kernel always performs better than using Gaussian atomic kernel. Indeed, histograms provide a rich distribution description of leaf attributes.

\begin{table}[!ht] 
\centering
\caption{Mean (and standard deviation) of overall accuracies computed over 100 repetitions for the artificial dataset. Best results (with a statistical significance less than 0.01\% considering the Wilcoxon signed-rank test for matched samples) between the rooted and subpath kernel -- Gaussian or $\chi^{2}$-- are boldfaced.} 
\begin{tabular}{ | l | m{1.6cm} | p{1.6cm} | p{1.6cm} | p{1.6cm} |}
	\hline
	\multicolumn{1}{|c|}{ \multirow{2}{*}{Scenario } } & \multicolumn{2}{c|}{ Rooted kernel} & \multicolumn{2}{c|}{ Subpath kernel}\\ \cline{2-5}
	&  \multicolumn{1}{c|}{Gaussian  } & \multicolumn{1}{c|}{$\chi^{2}$  } & \multicolumn{1}{c|}{Gaussian} & \multicolumn{1}{c|}{$\chi^{2}$  }  \\ 
	\hline
	(a) discriminative root only & \textbf{100.0} (0.0) & \textbf{100.0} (0.0) & \textbf{100.0} (0.0) & \textbf{100.0} (0.0) \\
	\hline
	(b) discriminative structure only & 48.9 (4.6) & 49.4 (3.4)  & \textbf{100.0} (0.0)  & \textbf{100.0} (0.0)  \\
	\hline
	(c) discriminative nodes only &  49.5 (3.5) & 51.2 (4.6) & \textbf{100.0} (0.0)  & \textbf{100.0} (0.0)  \\
	\hline

\end{tabular}
\label{tab:toy}
\end{table}

\subsection{Satellite Image Dataset}

\begin{figure}[!ht]
\centering
\includegraphics[height=3.25cm]{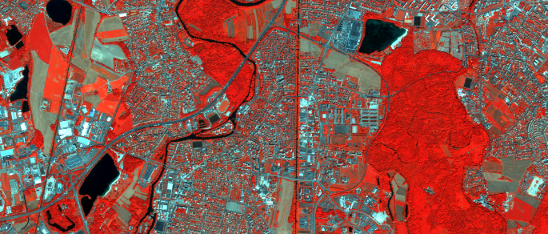}\hfill
\includegraphics[height=3.25cm]{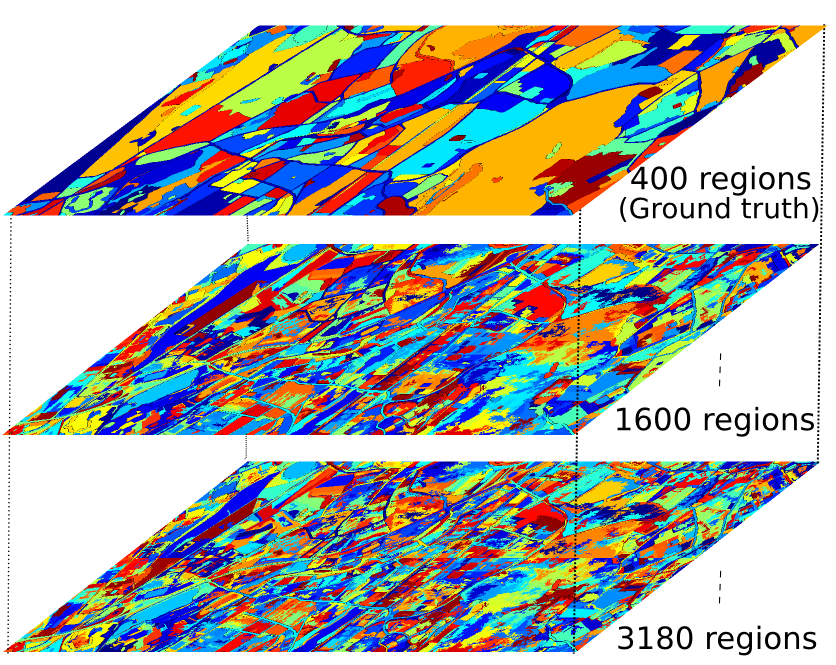}
\caption{\textsc{Strasbourg} dataset. Left: Color composition of a multispectral Quickbird image, \copyright DigitalGlobe, Inc. Right: examples of RHSEG segmentations at different scales (top: associated ground truth map with 400 regions).}
\label{fig:dataset} 
\end{figure}

Beyond the evaluation conducted on some artificial datasets, we also perform some experiments on real datasets. Since hierarchical representations are common in remote sensing, we explore the relevance of the proposed kernel in this domain. To do so, we consider a \textsc{QuickBird} satellite image with high spatial resolution (\textit{i.e.}, $2.4$ m per pixel) of Strasbourg Illkirch in France, initially proposed and discussed in \cite{kurtz2012extraction}. We can perform quantitative evaluation of the kernel-based classification procedure thanks to the availability of a ground truth (a partition of the initial image into 400 regions, each of them being associated to one of the 7 classes of interest, see list and distribution in Fig.~\ref{fig:rsresults}). Figure \ref{fig:dataset} shows the satellite image and its associated ground truth.

We compute a tree representation of each single region of the ground truth, using a standard open source hierarchical image segmentation method called RHSEG \cite{tilton2010rhseg}. RHSEG allows us to produce a fine segmentation map containing 3180 regions, that are subsequently aggregated to build coarser layers or segmentation maps with less regions (each iteration contains 300 regions less), as shown in Fig.~\ref{fig:dataset}. The coarsest segmentation is nothing but the ground truth. These different layers are stacked within tree structures. The last step consists in deleting the redundant nodes that remain unchanged through different scales. Finally we obtain 400 different trees, where each root represents a ground truth region and other nodes represent its components (subregions) at different scales. 

\subsubsection{Results}
\begin{figure}[!ht]
\center
\includegraphics[width=.9\textwidth]{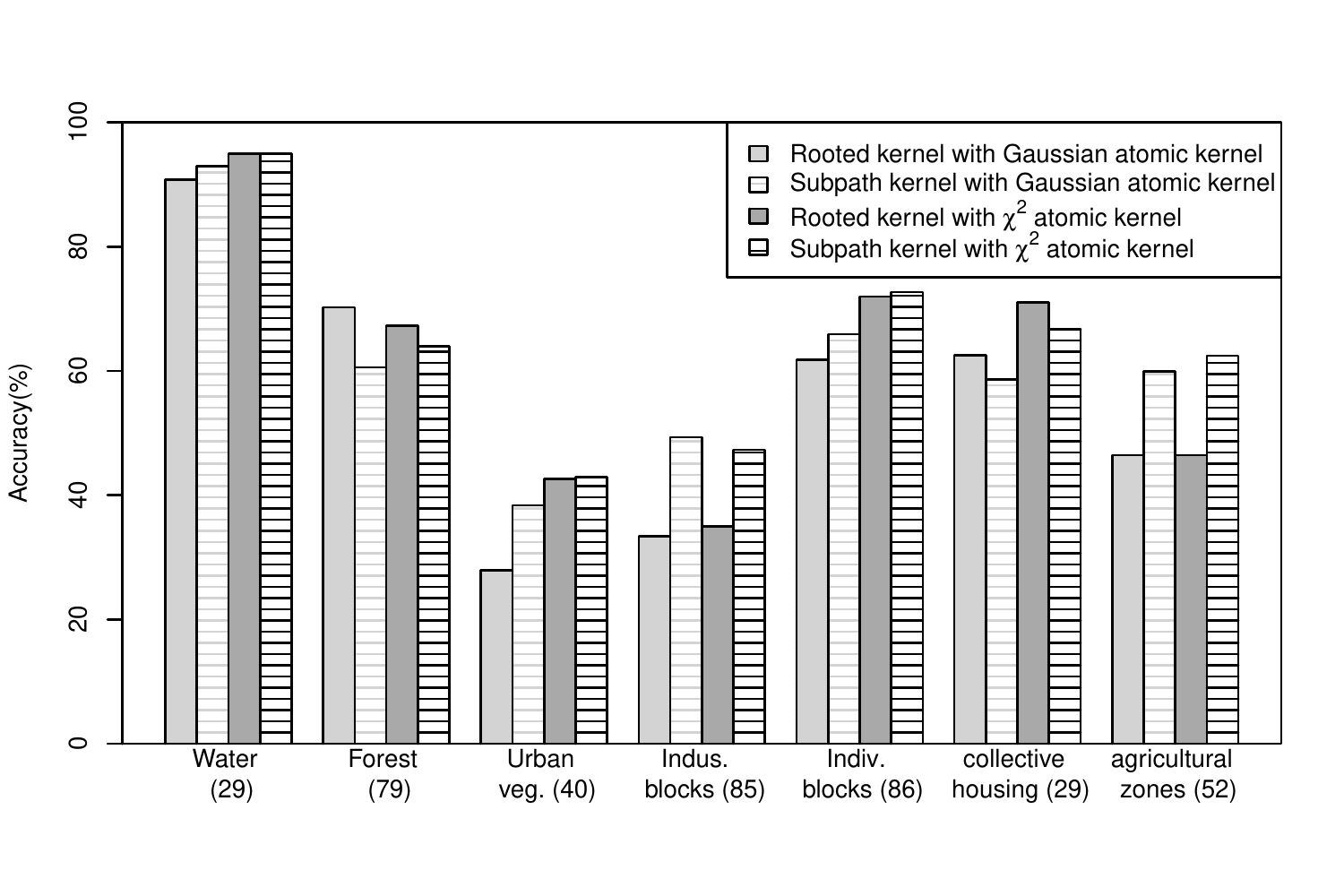}
\caption{Classification accuracy for the 7 different classes (number of samples or regions in the ground truth is provided in parentheses).}\label{fig:rsresults}
\end{figure}

Both rooted and subpath kernels are involved in a supervised classification process. Several statistics are derived: overall accuracy (ratio of correctly classified regions), average accuracy (average of the accuracy measured on each class), and kappa index (percentage of agreement in the test set, corrected by the agreement that could be expected by chance). Results are reported in Tab.~\ref{tab:resutls}. We can see that the proposed subpath kernel always outperforms rooted kernel. It is able to exploit the additional spatial features provided by the hierarchical representation of individual regions. A deeper analysis is provided in Fig.~\ref{fig:rsresults} where accuracies are provided for each class. Subpath superiority is mainly observed on classes such as urban vegetation, industrial urban blocks, and agricultural zones. For some other classes, performances are weaker. Let us note that the reported results are for the trade-off $\beta$ parameter providing the best overall accuracies. It may then lead to non adequate values for some classes, as by definition of the proposed kernel, setting $\beta=\infty$ would mimic the rooted kernel.

\begin{table}[!ht]
\centering
\caption{Mean (and standard deviation) of overall accuracies (OA), average accuracies (AA) and Kappa statistics ($\kappa$) computed over 100 repetitions for Strasbourg dataset. Computation time (in seconds) is also reported. Best results (with a statistical significance less than 0.01\% considering the Wilcoxon signed-rank test for matched samples) are boldfaced.} 
\begin{tabular}{|l|c|c|c|c|}
	\hline
	 \multicolumn{1}{|c|}{Method}& OA$[\%]$ & AA$[\%]$ & $\kappa$ & time  \\ 
	\hline
	Rooted kernel, with Gaussian atomic kernel& 53.1  (3.0) & 56.2 (2.9) & 0.447 (0.03) &  1.4 \\
	Subpath kernel, with Gaussian atomic kernel& \textbf{58.4} (2.6)	& \textbf{60.8} (2.9) & \textbf{0.498} (0.03) & 19.5 \\
	\hline
	Rooted kernel, with $\chi^{2}$ atomic kernel& 57.8 (2.2) & 61.3 (2.6) & 0.494 (0.03) & 2.7  \\
	Subpath kernel, with $\chi^{2}$ atomic kernel& \textbf{61.4} (2.8) & \textbf{64.4} (2.9) & \textbf{0.532} (0.03) & 98.8 \\

	\hline
\end{tabular}
\label{tab:resutls}
\end{table}


\section{Conclusion}
In this paper, we  introduced a new structured kernel that is able to cope with unordered trees equipped with numeric data. By doing so, we were able to apply pattern recognition and machine learning techniques to hierarchical image representations that become more and more popular, especially in remote sensing. We built upon a subpath kernel initially designed for bioinformatics data, as well as some graph kernels that relies on random walks. We show by some preliminary experiments the abilities and robustness of the proposed kernel. The encouraging results call for further investigation.

Among future research directions, a comparison with existing kernels in image analysis is planned. Most of them are based on graph kernels. Since trees are a particular class of graphs, various graph kernels may be considered (e.g. \cite{dupe2009tree}).

\section*{Acknowledgements}
The authors acknowledge the support of the French Agence Nationale de la Recherche (ANR) under reference ANR-13-JS02-0005-01 (Asterix project), and the support of R\'egion Bretagne and Conseil G\'en\'eral du Morbihan (ARIA doctoral project).
The authors would also like to thank A.~Puissant from LIVE UMR CNRS 7362 (University of Strasbourg) for providing the Strasbourg dataset (Quickbird image and ground truth).

\bibliographystyle{splncs03}
\bibliography{reference}
\printindex

\end{document}